\documentclass[journal]{IEEEtran}
\usepackage{cite}

\usepackage{amssymb}
\usepackage{graphicx}
\ifCLASSINFOpdf
\else
\fi
\usepackage[cmex10]{amsmath}
\usepackage{algorithmic}

\usepackage{array}
\usepackage{fixltx2e}
\usepackage{dblfloatfix}
\usepackage{url}

\usepackage{hyperref}
\usepackage{amsmath}
\usepackage{graphicx, subcaption}
\usepackage{setspace}
\usepackage{float}
\usepackage{cite}
\usepackage[british]{babel} 
\usepackage{float}
\usepackage[table]{xcolor}
\usepackage{color}
\usepackage{xcolor}
\usepackage{listings}
\usepackage{changepage}
\usepackage{algorithm}
\usepackage{algorithmic}
\usepackage{xr}
\usepackage{caption}
\usepackage{subcaption}
\usepackage{graphicx}
\usepackage{multirow}
\usepackage{amsfonts}

\usepackage{cleveref}

\begin{document}
%
\title{Synthetic Data Matters: Re-training with Geo-typical Synthetic Labels for Building Detection}
%
%
%

\author{Shuang~Song,~%
        Yang~Tang,~%
        and~Rongjun~Qin,~\IEEEmembership{Senior~Member,~IEEE}
\thanks{The authors are supported in part by the Office of Naval Research [grant numbers N000142012141 and N000142312670] and the Intelligence Advanced Research Projects Activity (IARPA) via Department of Interior/Interior Business Center (DOI/IBC) contract number 140D0423C0075.}
\thanks{\textit{Corresponding author: Rongjun Qin.}}
\thanks{S. Song is with the Geospatial Data Analytics Laboratory, the Department of Civil, Environmental and Geodetic Engineering, and the Translational Data Analytics Institute, The Ohio State University, Columbus, OH 43210 USA.}%
\thanks{Y. Tang is with the Geospatial Data Analytics Laboratory and the Department of Civil, Environmental and Geodetic Engineering, The Ohio State University, Columbus, OH 43210 USA.}
\thanks{R. Qin is with the Geospatial Data Analytics Laboratory, the Department of Civil, the Environment and Geodetic Engineering, the Department of Electrical and Computer Engineering, and the Translational Data Analytics Institute, The Ohio State University, Columbus, OH 43210 USA (e-mail:qin.324@osu.edu).}}

%
%

\markboth{Journal of \LaTeX\ Class Files,~Vol.~13, No.~9, September~2025}%
{Shell \MakeLowercase{\textit{et al.}}: Bare Demo of IEEEtran.cls for Journals}
%



\maketitle

\begin{abstract}

Deep learning has significantly advanced building segmentation in remote sensing, yet models struggle to generalize on data of diverse geographic regions due to variations in city layouts and the distribution of building types, sizes and locations. However, the amount of time-consuming annotated data for capturing worldwide diversity may never catch up with the demands of increasingly data-hungry models. Thus, we propose a novel approach: re-training models at test time using synthetic data tailored to the target region’s city layout. This method generates geo-typical synthetic data that closely replicates the urban structure of a target area by leveraging geospatial data such as street network from OpenStreetMap. Using procedural modeling and physics-based rendering, very high-resolution synthetic images are created, incorporating domain randomization in building shapes, materials, and environmental illumination. This enables the generation of virtually unlimited training samples that maintain the essential characteristics of the target environment. To overcome synthetic-to-real domain gaps, our approach integrates geo-typical data into an adversarial domain adaptation framework for building segmentation. Experiments demonstrate significant performance enhancements, with median improvements of up to 12\%, depending on the domain gap. This scalable and cost-effective method blends partial geographic knowledge with synthetic imagery, providing a promising solution to the “model collapse” issue in purely synthetic datasets. It offers a practical pathway to improving generalization in remote sensing building segmentation without extensive real-world annotations.
\url{https://github.com/GDAOSU/geotypical_synthetic_label_building_detection} 
\end{abstract}


%
\IEEEpeerreviewmaketitle

\section{Introduction}\label{sec:introduction}

\IEEEPARstart{A}{ccurate} global building segmentation from very high-resolution (VHR) satellite imagery is essential for applications like urban planning, disaster management, humanitarian assistance, and population estimation. However, this task is challenging due to the complex and diverse urban patterns across different geographical regions. Existing public datasets such as INRIA Aerial Image Labeling Dataset \cite{maggiori_can_2017}, Defence Science and Technology Laboratory (DSTL) Dataset \cite{benjamin_DSTL_2016}, and SpaceNet \cite{van_etten_spacenet_2019} provide large-scale datasets, but the coverage remains limited relative to global diversity. This limitation restricts model generalization across different regions, as variations in building types, urban layouts \cite{dangDifferentUrbanizationLevels2023}, and green spaces \cite{kong_synthinel-1_2020} introduce significant challenges.

\begin{figure}[!t]
    \centering
    \includegraphics[width=\linewidth]{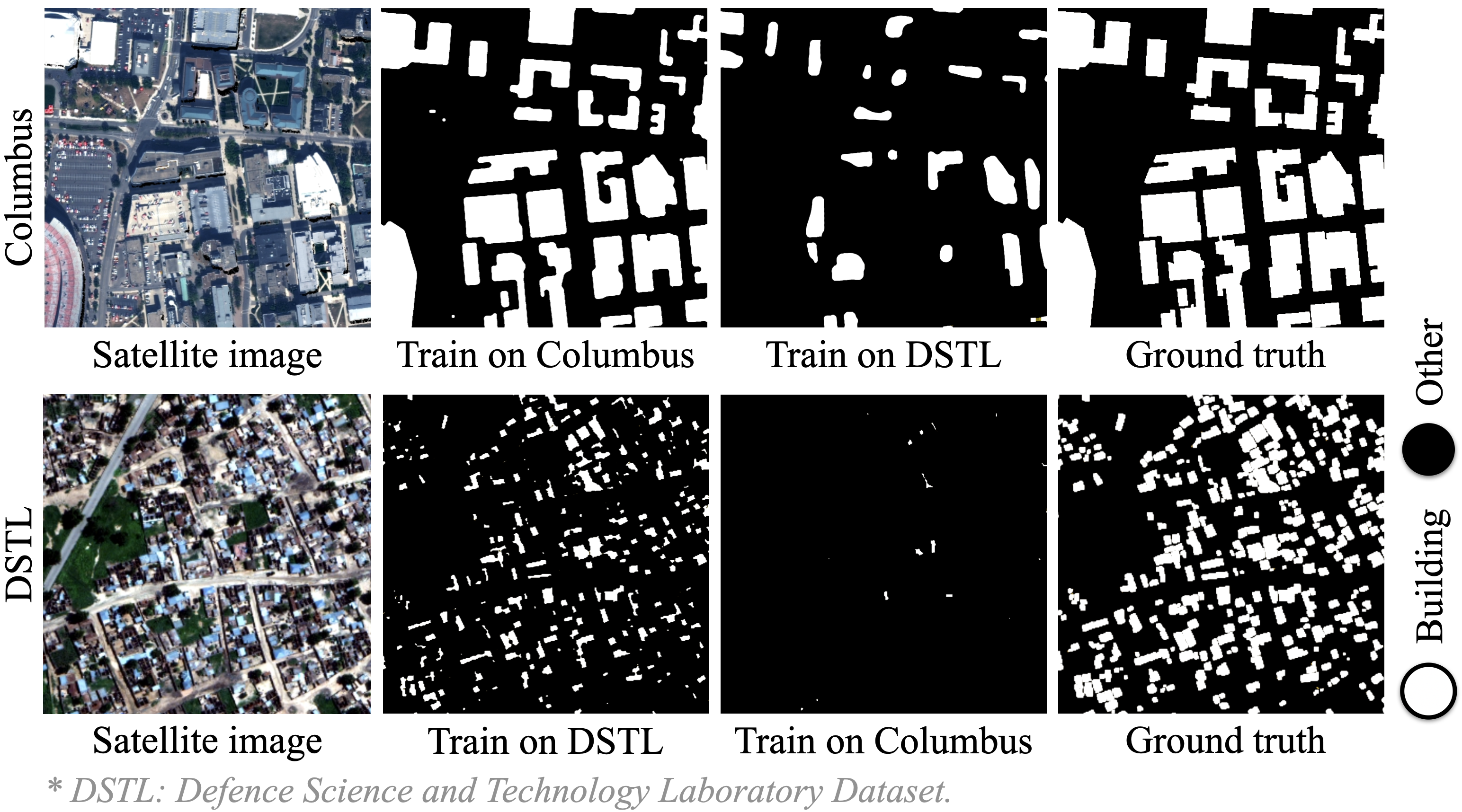}
    \caption{
    Cross-domain performance of building segmentation in very high-resolution satellite images. The top row shows results for Columbus, and the bottom row for the DSTL dataset \cite{benjamin_DSTL_2016}. Columns show the original image, predictions from same-domain and cross-domain models, followed by ground truth.}
    \label{fig:Problem}
\end{figure}

Generalization issues are evident when deep learning models trained on one dataset are applied to another with distinct characteristics. These challenges include performance degradation caused by differences in urban and rural distributions \cite{subediLeveragingNAIPImagery2023, wuLightweightConditionalConvolutional2023}, variations in city layouts, building densities, architectural styles \cite{sunMultiLevelPerceptualNetwork2023, sariturkComparativeAnalysisDifferent2023, liLightweightParallelOctave2023, ahmadCellularAutomataApproach2023}, and domain gaps between datasets that significantly impact accuracy \cite{zhang2022transformer, yan2022foreground, yu2021msegnet, li2022feature}. For instance, as shown in Figure \ref{fig:Problem}, models trained on the Columbus dataset perform poorly on the DSTL dataset, and vice versa, due to differences in building sizes, textures, layouts, and distributions.

Current approaches to addressing domain gaps include data augmentation (DAug) \cite{shorten_survey_2019, alomar_data_2023}, domain adaptation (DA) \cite{tzeng_adversarial_2017, huang_auggan_2018}, and few-shot learning. DAug applies transformations to enhance robustness but only modifies existing data, adding bias without introducing new variance. DA aligns source and target domain distributions using target data but often increases bias without fully addressing domain-specific diversity. Few-shot learning incorporates limited target samples to support DA but is constrained by data availability. Recent works combine these approaches with synthetic datasets \cite{richter_playing_2016, ros_synthia_2016, kong_synthinel-1_2020} to expand training data. However, models trained on data from a single city with fixed layout patterns, such as those generated by systems like MatrixCity \cite{li2023matrixcity}, still face significant generalization challenges.



This paper presents a scalable and cost-effective solution to address the challenges in building segmentation. Leveraging open-source data sources such as global road networks and building footprints from OpenStreetMap \cite{OpenStreetMap}, combined with physics-based rendering \cite{blender_online_community_blender_2021,epic_games_unreal_2019,haas_history_2014}, we generate geo-typical synthetic VHR satellite images and annotations. These synthetic datasets are specifically tailored to replicate the diversity and complexity of real-world urban layouts, enabling effective and targeted segmentation in regions with limited or incomplete data availability.
\begin{itemize}
    \item We propose a novel approach for generating synthetic data that replicates the geo-typical and architectural characteristics of target regions.

    \item We introduce a moderated domain randomization method to enhance the appearance of the synthetic data without introducing extra domain gaps by blind randomization.

    \item We present a comprehensive building segmentation framework that utilizes the adversarial domain adaption (ADA) to combine geo-typical synthetic data and high-quality annotations from other domains.

    \item Experiments highlight the utility and effectiveness of the proposed approach, achieving significant improvements in building segmentation performance, with gains of up to 12\% and, in certain cases, up to 29\%. 
\end{itemize}

The remainder of this paper is organized as follows. Section \ref{sec:related_works} reviews related works on synthetic datasets, DA for semantic segmentation. Section \ref{sec:methodology} describes our methodology for the generation of synthetic data. Section \ref{sec:experiment} presents performance evaluations and ablation study, and Section \ref{sec:conclusion} concludes the paper.

\section{Related Works}\label{sec:related_works}

\textbf{Building segmentation in remote sensing} has been widely studied using both classical and deep learning techniques. Traditional methods typically rely on handcrafted features and thresholding \cite{gu2018building, liu2022ndbsi} or object-based image analysis \cite{blaschke2010object}, but struggle with generalization across scenes. Deep learning-based methods \cite{luo2021deep, gui2022sat2lod2}, particularly convolutional neural networks such as U-Net \cite{ronneberger2015u} and DeepLab \cite{chen2017deeplab}, have substantially improved performance by learning spatial features directly from annotated data. Recent advances incorporate attention mechanisms and transformer-based backbones (e.g., RefineNet \cite{lin2017refinenet}, MSegNet \cite{yu2021msegnet}, Swin-Transformer \cite{liu2021swin}) to better capture long-range dependencies and contextual information.

Foundation models such as Segment Anything Model (SAM) \cite{kirillov2023segment} have shown promising generalization in natural images, their direct application to remote sensing tasks remains limited. Although efforts like SAMRS \cite{wang2023samrs} demonstrate potential through adaptation, they still rely on partial supervision or transfer learning rather than large-scale retraining from remote sensing labels. Despite these architectural developments, building segmentation in remote sensing remains fundamentally constrained by data limitations. Many existing benchmark datasets (e.g., INRIA \cite{maggiori_can_2017}, SpaceNet \cite{van_etten_spacenet_2019}, DSTL \cite{benjamin_DSTL_2016}) lack sufficient geographic diversity, coverage across varying building types and environments and sensor modalities. This presents significant challenges, particularly for training data-intensive models. These challenges motivate the exploration of domain adaptation and synthetic augmentation approaches, as proposed in this work.

\textbf{Synthetic dataset generation} has recently received significant attention. The main idea is to generate realistic-looking synthetic data with high-quality labels for training machine learning models, which can reduce the need for expensive and time-consuming data collection and annotation. 


Creating synthetic datasets for remote sensing is challenging due to the extensive scene coverage required and the shortage of public high-quality large-scale 3D urban landscape models. Synthinel-1 \cite{kong_synthinel-1_2020} provides a synthetic dataset for building detection, combining nine predefined city layouts for dataset creation. It proves that synthetic data can improve generalization capability. SyntCities \cite{reyes2022syntcities} is designed for disparity estimation, the task is less geolocation sensitive but regarding the land use and land cover tasks, the diversity of the dataset is not enough. SynRS3D \cite{song2024synrs3d} offers the so-called largest synthetic dataset for land cover mapping and height estimation, while SyntheWorld \cite{song2024syntheworld} focuses on land cover mapping and building change detection. Both works utilize GPT \cite{radford2021learning} and Stable Diffusion \cite{rombach2022high} for texture asset generation. The generative model enlarged the diversity, however, it may introduce domain gaps and bias of the pre-trained model, making it less effective for a specific target domain. While SyntheWorld and SynRS3D represent the current SOTA of high-diversity RS synthetic data using large-scale generative models, our approach complements them by embedding real-world geospatial constraints into procedural modeling. This ensures geographic consistency and enhances transferability, as validated by comparative results in Section \ref{sec:experiment}.

\textbf{Domain adaptation} aims to bridge domain gaps for machine learning models where source and target domains differ in data or feature distributions. Common DA methods include instance re-weighting \cite{9554236,10375080,7866898}, self-training \cite{9516689,9645575}, zero-shot learning \cite{LI2021145,9321719}, and adversarial learning \cite{9857935,CHEN2023169}. 

Adversarial domain adaptation (ADA), a subset of adversarial learning \cite{tzeng_adversarial_2017}, is widely adopted for aligning data distributions. ADA trains a discriminator to distinguish between source and target domain features while the segmentation model generates features indistinguishable to the discriminator. This approach effectively reduces domain gaps but primarily focuses on feature-level alignment, which may struggle with diverse or evolving data distributions. Tzeng et al. \cite{tzeng_adversarial_2017} reported a 13.3\% accuracy improvement using ADA between Amazon product photo and webcam domains, demonstrating its efficacy in static domain scenarios. However, ADA often fails in dynamic or multi-source settings, where evolving data distributions and diverse domain characteristics require greater flexibility. To address this, DAugNet \cite{tasar2020daugnet} was developed to align input data across domains. It introduces a lifelong unsupervised approach tailored for satellite imagery, enabling multi-source and multi-target domain adaptation. By leveraging adversarial losses to transfer styles and adding domain-specific layers for new distributions, DAugNet continuously adapts to unseen data while retaining performance on previously learned domains. While DAugNet focuses on dynamic and input-level alignment, it does not explicitly handle semantic inconsistencies. To address this, Category-level Adversarial Network (CLAN) \cite{luo_category-level_2022, luo_taking_2019} was introduced, emphasizing category-level alignment. By embedding purified features and enabling category-level adversarial learning, CLAN reduces semantic inconsistencies and enhances domain calibration. Building on these advances, Chen et al. \cite{chen2022unsupervised} proposed a method combining global and category-level adversarial learning with a category-certainty attention module to improve alignment in poorly matched regions. This approach enhances both global consistency and local semantic alignment, particularly in VHR satellite imagery. It outperforms traditional methods like AdaptSegNet \cite{tsai_learning_2018} and CLAN \cite{luo_category-level_2022}, demonstrating superior segmentation accuracy in challenging domain gaps.


\section{Methodology}\label{sec:methodology}
\begin{figure*}[h]
    \centering
    \includegraphics[width=\linewidth]{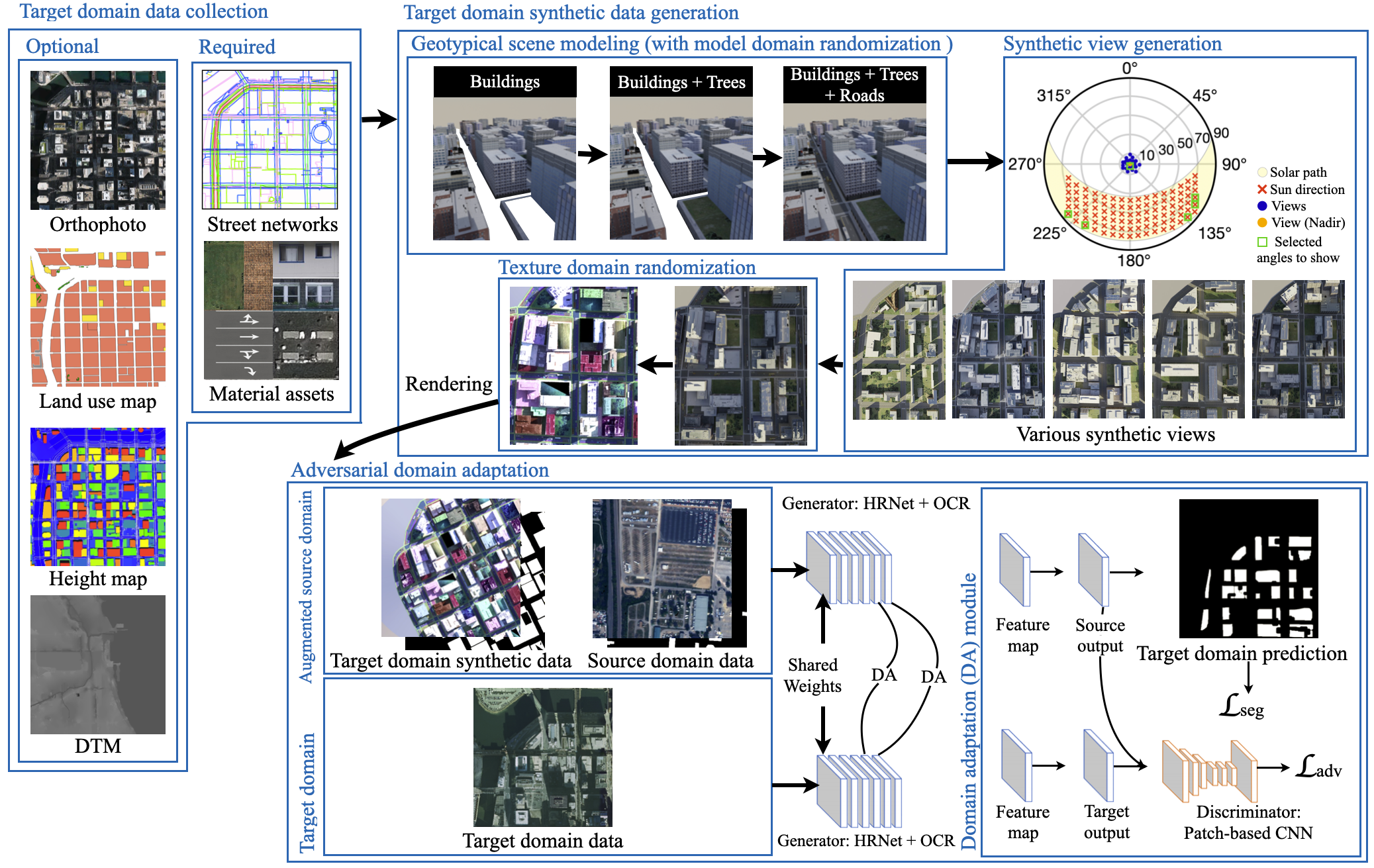}
    \caption{Schematic workflow of our proposed method, which illustrates the use of target-specific synthetic data for enhancing building segmentation. Note, $M^\star$ is an optional material. DTM refers to Digital Terrain Model, in our case we used Shuttle Radar Topography Mission (SRTM) digital elevation data \cite{farr2007shuttle}.} 
    \label{fig:overview}
\end{figure*}

Typical synthetic data generation leverages 3D assets to create diverse views and modalities \cite{ros_synthia_2016,richter_playing_2016,shah_airsim_2018}, addressing limitations such as insufficient labeled data for tasks like stereo matching, height estimation, and semantic labeling. Our approach builds on this concept by specifically tailoring datasets to match the city layout distributions of target domains. Using auxiliary data from OpenStreetMap, we procedurally generate scenes with randomized 3D objects, including buildings and trees, aligned with street network patterns. These scenes are rendered as synthetic VHR satellite images with corresponding semantic labels, ensuring the generated data aligns closely with the geographic characteristics of the target domain.

The overall workflow is rather straightforward and depicted in Figure \ref{fig:overview}, which consists of three major steps: 1. Target domain data collection: This section outlines the data requirements and their importance. 2. Target Domain Synthetic Data Generation: This section introduces the procedures for model generation, model rendering, and texture domain randomization (DR). 3. Adversarial domain adaptation: Presents the framework utilized and the rationale behind its selection. The following subsections detail the workflow of our methodology.

\subsection{Target domain data collection} \label{subsec:data_collection}

The data used for generating synthetic scenes focuses on buildings, trees, and roads. The street network, composed of nodes (intersections) and edges (streets), serves as the primary input, providing road segment lengths and categories for deriving road widths. It also forms the foundation for generating building blocks and green areas, defining the city's layout. In the absence of a land use map, the street network can be analyzed to classify areas as residential or commercial based on weighted aggregation of road categories (e.g., primary, residential, trunk).



When height maps are unavailable, land use types (e.g., residential or commercial) can estimate building heights by sampling from category-specific ranges. In U.S. cities, significant height differences between commercial and residential buildings enable effective random sampling, whereas smaller gaps in Asian and European cities favor height map usage. Alternatively, OpenStreetMap data, such as floor numbers, can be converted into height estimates (e.g., assuming 3 meters per floor), offering a practical solution for diverse urban environments.

Another required dataset is Material assets, which are essential for defining building styles and shaping synthetic city appearances. Our approach is sourced pre-built materials from ESRI. Additional textures, such as orthophotos from NAIP\footnote{\url{https://naip-usdaonline.hub.arcgis.com/}}, enhance realism. Inspired by SyntheWorld \cite{song2024syntheworld}, stable diffusion with GPT-generated prompts can also generate textures.


Terrain elevation data (DTM) is optional, it enhances the alignment of the synthetic model to the target domain. In our experiments, we used the Shuttle Radar Topography Mission (SRTM) dataset \cite{farr2007shuttle}, which is sufficient enough, though higher-resolution regional DTMs could further improve accuracy and realism.


\subsection{Target domain synthetic data generation} \label{subsec:data_generation}
After collecting the target domain data, the focus of this section shifts to target domain synthetic data generation. This process is designed to create diverse and realistic datasets tailored to the characteristics of specific urban environments. This section outlines three key components: 1. Geo-typical scene modeling involves creating synthetic buildings, trees, and roads using procedural methods informed by the data outlined in Section \ref{subsec:data_collection}. 2. Synthetic view generation section includes rendering scenes under varying lighting conditions. 3. Texture DR details our approach that balances domain alignment and enhances robustness without increasing the domain gap.

\subsubsection{Geo-typical scene modeling} \label{subsubsec:scene_modeling}

The goal is to ensure geographic realism and improve robustness and accuracy in analysis by aligning synthetic elements with target domain urban structures. Using the street network as a key constraint, locations and areas of roads, buildings, and trees are determined to match the underlying city layout. While locations remain fixed, model DR varies the appearance and attributes, such as building heights, materials, and tree densities, within distributions reflective of land use types.


Our city model is constructed by combining three components: buildings, trees, and roads. The union of these components forms the complete city model (\ref{equ:0}). Models for different land use types are generated collectively, as they share similar procedural parameters.


\begin{equation}
\begin{aligned}
\text{CityModel} = 
\{ \mathcal{B}_l \cup \mathcal{T}_l \cup \mathcal{R} \mid l \in L\},
\end{aligned}
\label{equ:0}
\end{equation}

\noindent where \( \mathcal{B}_l \), \( \mathcal{T}_l \), and \( \mathcal{R} \) represent the building models, tree models, and road models, respectively, of land use type \( l \), and the set of all land use types denoted as \( L \). The street network $S$ divides the city into enclosed plots, which are further subdivided into polygonal-shaped lots. 

\textbf{Plot subdivision} is based on land use type, which defines minimum and maximum lot area constraints and defines Green Area Ratio (GAR). The GAR was determined by city zoning regulations, which specifies the required proportion of green area. To allocate green areas, the lots are sorted in ascending order and accumulated until the GAR requirement is met. The selected smaller lots are designated as green areas for tree generation, while the remaining lots are assigned for building generation.


\textbf{Building generation} begins with plot denoted as \( i \in P_l \), where $P_l$ is a set of all plots of land use type \( l \) and \( B_i \) represents the attributes of buildings in the plot as defined in \eqref{equ:B_l}.

\begin{equation}
\mathcal{B}_l = \{ B_i \mid i \in P_l\} .
\label{equ:B_l}
\end{equation}

\noindent Buildings in $B_i$ share a set of random variables as defined in \eqref{equ:B_i}. The attributes of individual buildings will be sampled from the given distributions and a 3D model will be constructed by CGA (Computer Generated Architecture) builder \footnote{https://doc.arcgis.com/en/cityengine/latest/help/help-cga-modeling-overview.htm}. 

\begin{equation}
B_{i} = \{ h_{i}, m_{i}, f_{i}, F_{i} \} .
\label{equ:B_i}
\end{equation}

\noindent The building height \( h_{i} \) is a scaled discrete uniform distribution ($\lfloor U \rfloor$) with three parameters: level height $H_{\text{level}}^l$, and range of levels 
$(\Gamma_{\text{min}}^l, \Gamma_{\text{max}}^l)$, as shown in \eqref{equ:building_height}. These parameters are derived from OpenStreetMap building attributes that intersect with the plot. When DTM is provided, the base height is determined by the elevation from DTM at the location to ensure the proper terrain attachment.

\begin{equation}
h_{i} = H_{\text{level}}^i * \lfloor U \rfloor(\Gamma_{\text{min}}^i, \Gamma_{\text{max}}^i) .
\label{equ:building_height}
\end{equation}

The building material subset \( m_{i} \) is drawn from material library \( M_{l} \) according to the land use type, reflecting the distinct appearance of different building types (e.g., residential and commercial buildings). Building footprints \( f_{i} \) are generated during plot subdivision, with specific lots designated as building footprints. The subdivision behavior varies by land use type; for instance, commercial buildings typically occupy larger areas. Additional features \( F_{i} \) define specific characteristics of the building model, such as roof type, stairs, chimneys, and curtain walls. These features are assigned according to the land use type. 


\textbf{Tree generation} is performed for each land use type \( l \) by identifying green areas and tree models \( \mathcal{T} \). The tree attributes for lot green areas in set of plots \( i \in P_l \) are defined in (\ref{equ:5}):

\begin{equation}
\mathcal{T}_l = \{ T_i \mid i \in P_l \}.
\label{equ:5}
\end{equation}

Tree models \( T_i \) represent the tree models in green areas \( i \). The number of trees is determined based on these areas and is randomly assigned following the Poisson disk sampling \cite{bridson2007fast}. Tree features, such as species and height, are specified based on the characteristics of the land use type \( l \).

\textbf{Road Generation} involves utilizing the street network \( S \) to extract segments, such as edges \( S_{\text{edge}} \) to form streets and sidewalks, and nodes \( S_{\text{node}} \) to generate intersections. The road model \( \mathcal{R} \) is defined in (\ref{equ:6}):

\begin{equation}
\mathcal{R} = \{ R_i, \Psi_j \mid i \in S_{\text{edge}}, j \in S_{\text{node}} \}.
\label{equ:6}
\end{equation}

Let \( R_i \) denote a road segment and \( \Psi_j \) denote a road intersection. Each road segment has parameters determined by the road class (\( t_i \)), such as width and the presence of sidewalks. The street network \( S \) is utilized to compute intersection angles and the degree of node at intersection \( j \). Based on these characteristics, intersection types (e.g., crossing, junction, roundabout) are classified. 

\subsubsection{Synthetic view generation} \label{subsubsec:view_generation}

For city-scale rendering, we use Lambertian BRDF because most buildings and environments observed from satellites are diffusely reflecting surfaces. This approach also saves computational resources and time.

The relationship between color and reflectance in the context of BRDF can be expressed by extending the BRDF and lighting integral formulas (\ref{equ:7}) to account for different color channels:
\begin{equation}
L_{r,\text{color}}(\omega_r) = \frac{\rho_{\text{color}}}{\pi} \int_{\Omega} L_{i,\text{color}}(\omega_i) \max(0, \omega_i \cdot n) \, d\omega_i,
\label{equ:7}
\end{equation}

\noindent where, \(\text{color}\) represents the specific color channel. \(L_{r,\text{color}}(\omega_r)\) is the reflected radiance in the direction \(\omega_r\) for a specific color channel. \(\rho_{\text{color}}\) is the reflectance for the specific color channel, and \(\pi\) is a constant used to normalize the BRDF. \(\Omega\) represents all incoming directions within the hemisphere above the surface. \(L_{i,\text{color}}(\omega_i)\) is the incident radiance from direction \(\omega_i\). Which is the direction of the incoming light above the surface, and \(n\) is the normal vector to the surface. The term \(\omega_i \cdot n\) is the dot product of the incoming light direction and the surface normal, ensuring that only front-side contributions are considered. Finally, \(d\omega_i\) represents the differential solid angle element in the incoming direction. This formulation ensures accurate reflectance properties for each color channel. To simulate realistic city lighting conditions, the sun is randomly positioned along all possible solar paths within the target domain (Figure \ref{fig:overview}). 

In our approach, we use Blender Cycles \cite{blender_online_community_blender_2021} as the renderer. Blender is open-source 3D content creation software that supports rendering. For automatically setting camera poses, we use the virtual camera generator \cite{song_vis2mesh_2021} to generate camera parameters. The camera orientation is randomly set, with an azimuth angle of 0° to 360° and an off-nadir angle of 0° to 10°. Its height is adjusted to match the target domain's ground sample distance (GSD) of 0.3m. To ensure the entire scene appears sharp, the focus is set to infinity, and the depth of field is disabled. The model is then rendered into colored images and corresponding class annotations.

\begin{figure}
    \centering
    \includegraphics[width=\linewidth]{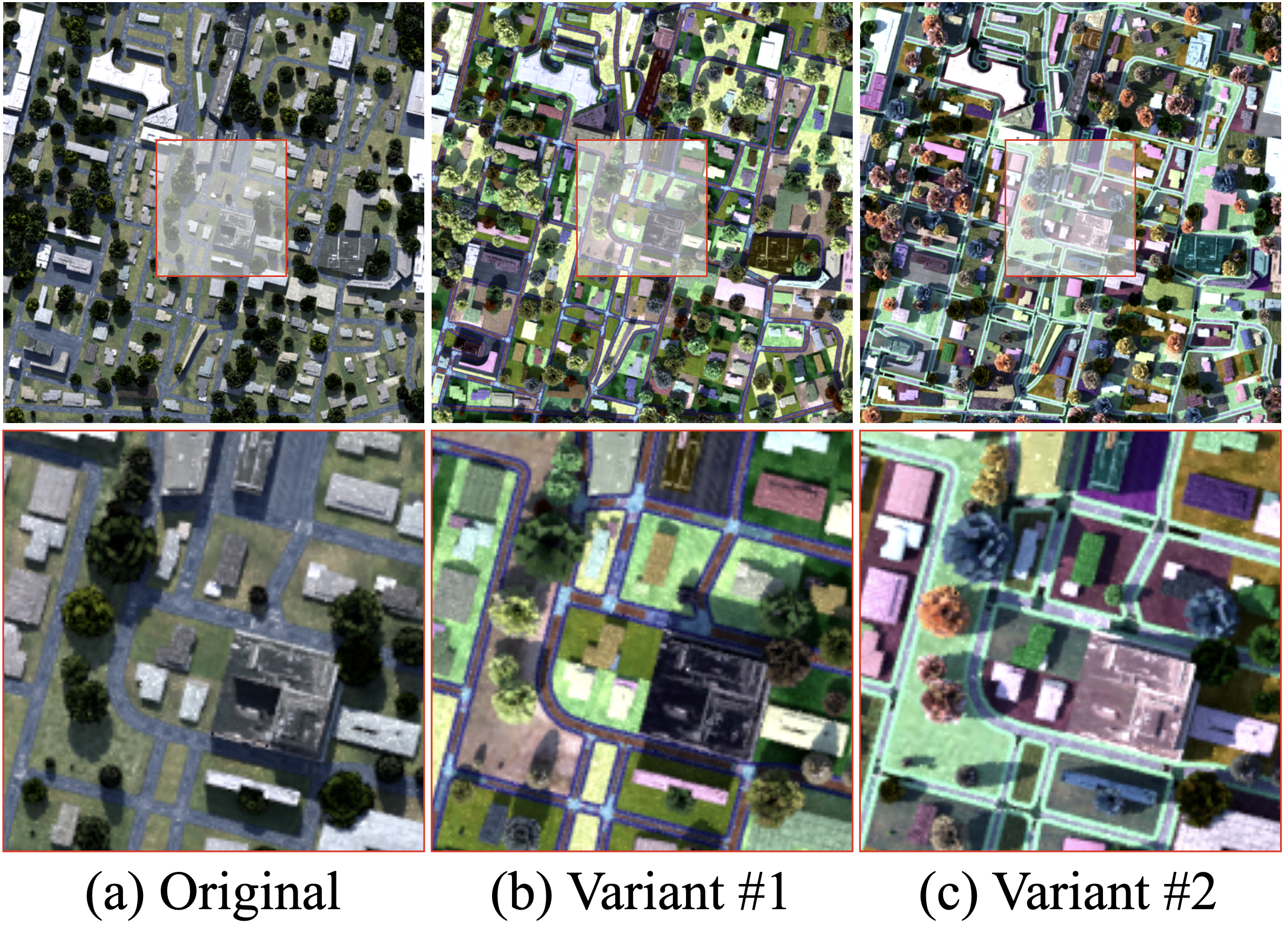}
    \caption{Sample images generated with texture domain randomization (DR)}
    \label{fig:domain_randomization}
\end{figure}

\subsubsection{Texture domain randomization} \label{subsubsec:texture_domain_random}

The goal of our approach is to enhance scene richness while avoiding introducing additional domain gaps. Typical DR methods in synthetic data generation randomize all objects in the scene to maximize diversity, but this blind randomization can easily increase domain gaps \cite{tobin2017domain, prakash2019structured}. To address this, we employ a selective DR technique that focuses on hue randomization in the texture process. Inspired by previous literature demonstrating that tailored domain randomization significantly outperforms standard augmentation techniques for cross-domain semantic segmentation tasks \cite{tobin2017domain, prakash2019structured}, our texture DR leverages procedurally created synthetic environments closely aligned with the structural characteristics of the target region. Additionally, our data augmentation approach is compatible with general data augmentation, which can be applied subsequently if necessary. This ensures realistic city structures are preserved while introducing necessary variability, as shown in Figure \ref{fig:domain_randomization}.


To introduce random variation to texture, we change the hue component of an image in the HSV color space. We first convert the RGB texture to HSV space, then shift the hue $H$ with a randomized value \(\Delta H\). The hue adjustment is defined as shown in (\ref{equ:8}):
\begin{equation}
H' = (H + \Delta H) \mod 360, \quad \Delta H \in [-180, 180],
\label{equ:8}
\end{equation}
\noindent the edited hue \(H'\) is then converted back to RGB space.


\subsection{Adversarial domain adaptation} \label{subsec:ADA}
Synthetic data generated in Section \ref{subsec:data_generation} introduce domain bias when compared to real-world data. For instance, synthetic datasets are less realistic and diverse, which could result in models less transferable to the real-world domain\cite{offenhuber2024shapes}. This domain gap can degrade model performance. ADA addresses this issue by aligning the feature distributions of the source and target domains, ensuring that the model learns domain-invariant features.

In this paper, we employ an unsupervised ADA framework called CLAN (Category-level Adversarial Networks) \cite{luo_category-level_2022,luo_taking_2019}. CLAN improves upon its predecessor, AdaptSegNet \cite{tsai_learning_2018}, by introducing a category-level adversarial learning mechanism. This mechanism aligns features at the category level, effectively reducing semantic inconsistencies between the source and target domains, which is crucial in tasks like semantic segmentation.

The generator in CLAN is originally implemented generator using ResNet101 \cite{he2016deep}, which is effective but limited in spatial precision. In our work, we replace the ResNet generator with HRNet + OCR \cite{yuan2020object} to enhance performance. HRNet excels in preserving high-resolution features and extracting multi-scale representations, while the OCR module (Object-Contextual Representations) further improves global context awareness, which is essential for accurate semantic segmentation. Ablation experiments comparing different backbones under the CLAN framework are provided in Section \ref{subsubsection4_Backbones} to support the performance claims.

The discriminator in CLAN is a patch-based fully convolutional network (FCDiscriminator) that progressively downsamples feature maps through strided convolutions, without skip connections or decoder layers. It is responsible for distinguishing between source and target domain features. The overall training framework is illustrated in Figure \ref{fig:overview}.

The ADA framework relies on two main components: a generator \(G\) and a discriminator \(D\). The generator \(G\) is responsible for extracting domain-invariant features from input images, leveraging its architecture to align feature distributions between the source and target domains. The discriminator \(D\), on the other hand, is tasked with distinguishing between features originating from the source and target domains. These two components are trained in an adversarial manner, where the generator aims to minimize the discriminator's ability to differentiate between the two domains, effectively learning domain-invariant features, as defined in (\ref{equ:9}). Meanwhile, the discriminator seeks to maximize its ability to correctly identify the domain of the input features, as defined in (\ref{equ:10}). This min-max training paradigm ensures the model achieves robust generalization across domains, reducing the performance gap caused by domain bias.

The segmentation loss function for generator $G$ is defined as:

\begin{equation}
    \mathcal{L}_{\text {seg}}(G)=E\left[\ell\left(G\left(X_S\right), Y_S\right)\right],
    \label{equ:9}
\end{equation}

\noindent where $E[\cdot]$ is statistical expression. $\ell(\cdot, \cdot)$ denotes a loss function, such as multi-class cross entropy, which used to train $G$ to predict pixel-level labels correctly. And The adversarial loss function is defined as:

\begin{equation}
    \begin{aligned}
    \mathcal{L}_{\text{adv}}(G, D)= & -\mathbb{E}\left[\log \left(D\left(G\left(X_S\right)\right)\right)\right] \\
    & -\mathbb{E}\left[\log \left(1-D\left(G\left(X_T\right)\right)\right)\right],
    \end{aligned}
    \label{equ:10}
\end{equation}

\noindent where $X_S$ and $X_T$ are the source and target domain images, respectively.

The framework is jointly optimized using two key loss functions: the segmentation loss \(\mathcal{L}_{\text{seg}}\) and the adversarial loss \(\mathcal{L}_{\text{adv}}\). The segmentation loss \(\mathcal{L}_{\text{seg}}\) supervises the generator \(G\) by ensuring that its predictions for the source domain align with the ground truth labels, thereby maintaining its ability to perform accurate semantic segmentation. Meanwhile, the adversarial loss \(\mathcal{L}_{\text{adv}}\) drives the adversarial training process between the generator \(G\) and the discriminator \(D\), aligning feature distributions from the source and target domains to reduce domain bias. Together, these losses enable the generator to produce domain-invariant features while preserving segmentation accuracy, ensuring robust generalization across domains. For more details please refer to Supplementary Section S1. 

While ADA and DR techniques aim to reduce domain gaps, they may introduce new gaps if not carefully constrained. Therefore, our framework incorporates moderated texture DR and category-level ADA to preserve structural integrity and semantic consistency. The ablation study results further confirm that these components improve generalization when domain differences are large, but may slightly impact performance when the source and target domains are already very close.

\section{Experiment}\label{sec:experiment}
\begin{table*}[h]
    \centering
    \begin{tabular}{c|c|c|c|c|c|c}
        \hline
        \textbf{Identifiers}& \textbf{Dataset} & \textbf{Patches} & \textbf{Bands} & \textbf{Sensor} & \textbf{GSD} & \textbf{Region and Scene} \\
        \hline
        \hline
        Cbus & Columbus & 1000 & 3 & WorldView & 0.3m & US Urban \\
        CHI & Chicago \cite{maggiori_can_2017} & 1200 & 3 & Aerial & 0.3m & US Dense Urban \\
        ATX & Austin \cite{maggiori_can_2017} & 1800 & 3 & Aerial & 0.3m & US Urban \\
        TIR & Tyrol-West \cite{maggiori_can_2017} & 1800 & 3 & Aerial & 0.3m & EU Sparse Urban \\
        DSTL & DSTL \cite{benjamin_DSTL_2016} & 1000 & 3 & WorldView & 0.31m & Various Places \\
        \hline
        Cbus\_{s} & Columbus Synthetic & 1200 & 3 & Synthetic & 0.3m & US Urban \\
        Cbus\_{sw} & Columbus Synthetic without DR & 101 & 3 & Synthetic & 0.3m & US Urban \\
        CHI\_{s} & Chicago Synthetic & 1200 & 3 & Synthetic & 0.3m & US Dense Urban \\
        SynRS3D & SynRS3D \cite{song2024synrs3d} & 1200 & 3 & Synthetic & 0.3m & Various Places \\
        SyntheWorld & Syntheworld \cite{song2024syntheworld} & 1200 & 3 & Synthetic & 0.3m & Various Places \\
        \hline
    \end{tabular}
    \caption{Characteristics and specifications of real-world and synthetic datasets used in the study}
    \label{tab:dataset_description}
\end{table*}

The experiment section addresses four aspects: dataset, training, metrics, and evaluation focus on two key objectives: 1. Evaluating the impact of synthetic data inclusion on model performance during transfer from real-world data to a target domain (Section \ref{section4_exp_real_vs_synthetic}). 2. Assessing the contribution of individual components, including ADA, texture DR, and city layout alignment, to the effectiveness of synthetic data (Section \ref{section4_ablation_study}).

\subsection{Dataset} \label{section4_dataset}

The effectiveness of our proposed approach is evaluated using ten datasets, shown in Table \ref{tab:dataset_description}, comprising five real-world datasets and five synthetic datasets. The real-world datasets include an in-house OSU dataset (Columbus, US), the INRIA aerial building dataset \cite{maggiori_can_2017} (Chicago, US; Austin, US; Tyrol-West, Vienna), and DSTL \cite{iglovikov2017satellite}. The synthetic datasets includes our method generated geo-typical synthetic data to replicate the city layouts of Columbus and Chicago, ensuring alignment with real-world urban structures. also includes the SOTA synthetic data SynRS3D \cite{song2024synrs3d}, SyntheWorld\cite{song2024syntheworld}, which employ large-scale generative models like GPT and Stable Diffusion to enhance texture diversity, that have demonstrated superior performance over prior synthetic benchmarks in their respective publications. These datasets, particularly the INRIA aerial building dataset and DSTL, are well-established and have been extensively used in prior studies, highlighting their diversity and reliability for image analysis tasks. By incorporating both real-world and synthetic datasets, our evaluation provides a comprehensive testbed to assess the generalization capabilities of the proposed method.

\begin{figure}[h]
    \centering
    \includegraphics[width=0.46\textwidth]{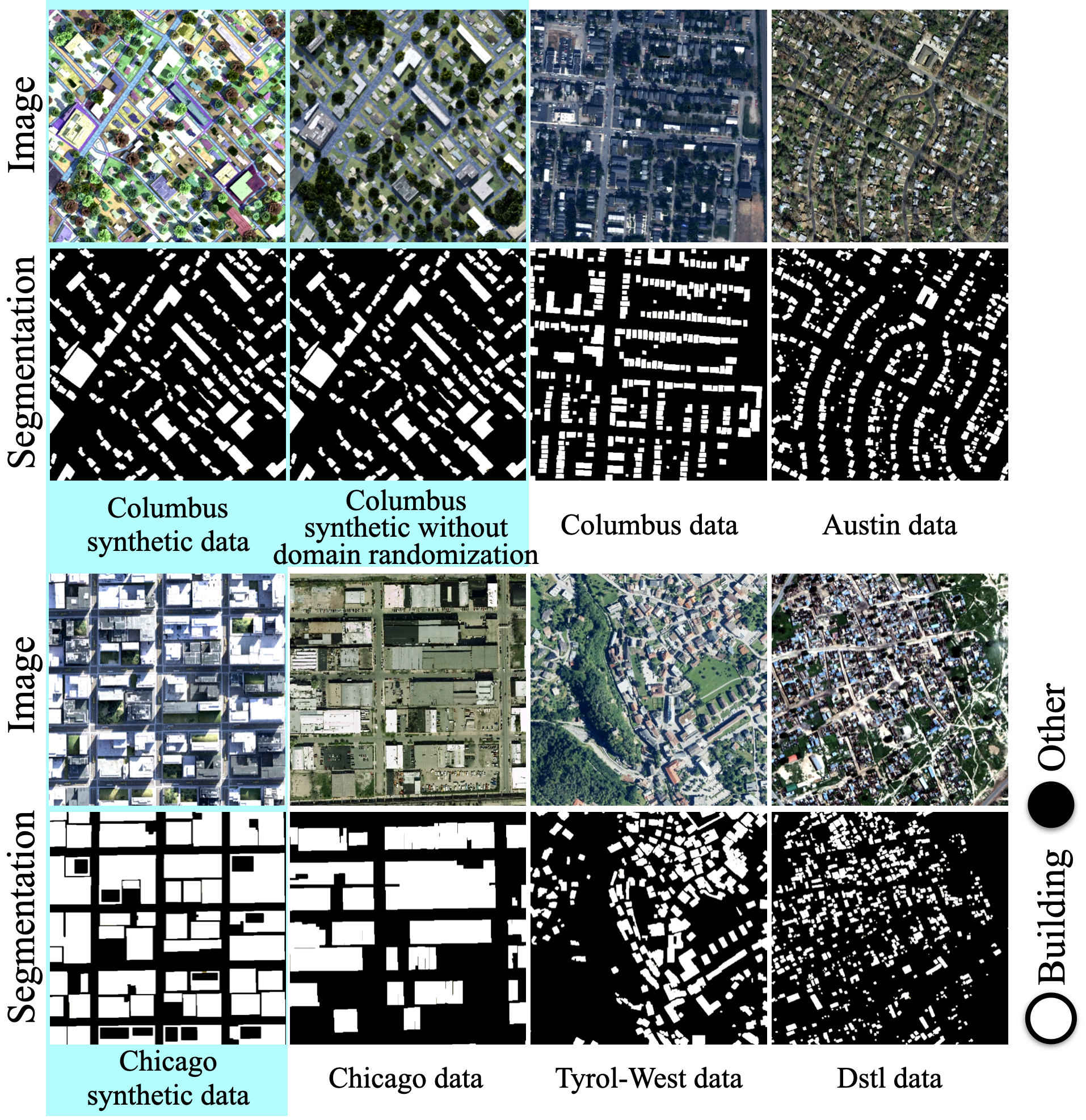}
    \caption{Example of real and our generated synthetic dataset. The cyan color in the background highlighted the synthetic data}
    \label{fig:fig_datasets}
\end{figure}

Samples of the eight datasets are illustrated in Figure \ref{fig:fig_datasets}, which represent a variety of urban patterns. From Columbus (Cbus), Chicago (CHI), Austin (ATX), Tyrol-West (TIR), and DSTL, as well as three synthetic datasets including synthetic Chicago (CHI\_{s}), synthetic Columbus (Cbus\_{s}) and synthetic Columbus data without texture DR (Cbus\_{sw}), which is a subset of (Cbus\_{s}). To facilitate easier reference and improve the clarity of our table layouts, we have assigned identifiers to each dataset. These identifiers, along with their corresponding dataset descriptions, are detailed in Table \ref{tab:dataset_description}.

Cbus and ATX are typical U.S. cities; the datasets include residential and commercial areas with moderately dense construction. CHI, with its dense urban architecture as America's third-largest city, the dataset includes large and compact buildings as well as residential areas, providing unseen scenes of urban environments. In contrast, the TIR dataset offers a distinctly European urban settlement with its architecture features sparse single houses and terraced commercial centers. The real-world data from CHI, ATX and TIR are sourced from the INRIA Aerial Image Labeling Dataset\cite{maggiori_can_2017}, which captured diverse illumination conditions due to different flight times. The DSTL dataset, originating from a Kaggle competition \cite{benjamin_DSTL_2016}, encompasses satellite imagery from multiple global locations, introducing a broad range of building scenarios. This variety ensures a robust dataset, adequately representing the complexity of urban environments. SynRS3D and SyntheWorld are included as they represent the SOTA and diverse synthetic datasets currently available. Since our study focuses on evaluating model generalization and robustness across globally urban environments, their inclusion provides a meaningful and relevant benchmark.

The evaluation focuses on a study scope of VHR building detection, hence all datasets presented in Table \ref{tab:dataset_description} are tailored for the building detection tasks. To align with our training objectives, we standardize the datasets to a 0.3m resolution and divide them into 512 $\times$ 512 pixel patches with a 50\% overlap. For model evaluation, we allocate 80\% of the data for training, reserving the remaining 20\% for validation.

We conducted 5 (coupled synthetic data) $\times$ 5 (source domain) $\times$ 2 (target domain) = 50 experiments to evaluate the model's performance across diverse geographic regions. The experiments were designed to include various urban patterns by utilizing combinations of real-world data from five urban environments with five synthetic datasets as source. Targeting two domains (Cbus and CHI) ensuring a sufficient range of scenarios for evaluation. For dataset integration during training, we employed a simple yet effective method: concatenating and shuffling patches from the source and synthetic datasets to ensure balanced exposure.

\begin{figure*}[h]
    \centering
    \includegraphics[width=0.88\textwidth]{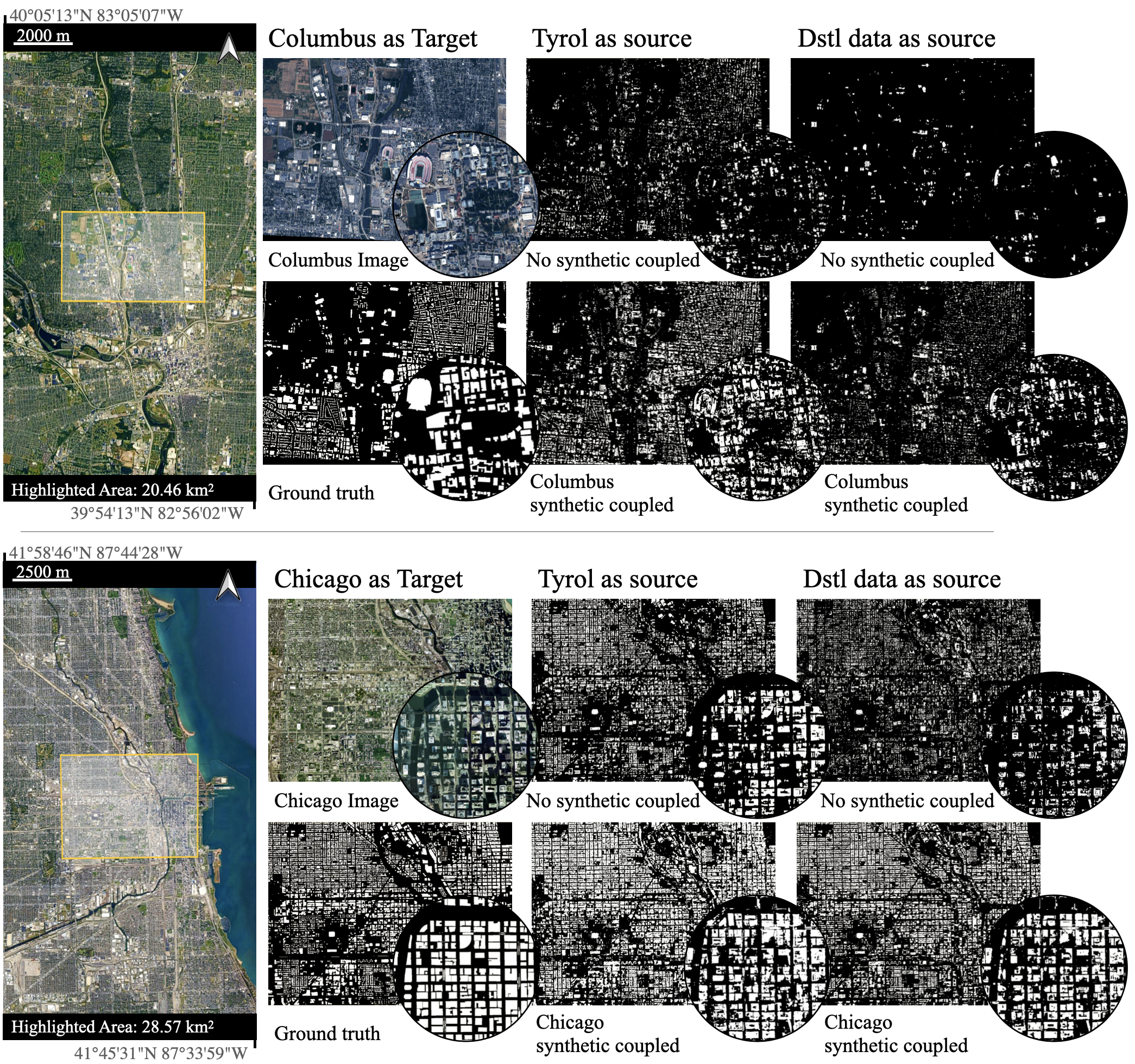}
    \caption{Comparative visualization of building segmentation predictions: Showcasing the impact of synthetic data coupling across different source domains}
    \label{fig:fig_results}
\end{figure*}

\subsection{Network training} \label{section4_training}
The training is conducted from scratch without any pretrained weights to avoid bias from unrelated domains. The generator is based on HRNet-W48 with an OCR head. The discriminator adopts a patch-based fully convolutional architecture (FCDiscriminator) with five convolutional layers and LeakyReLU activations, performing patch-level binary classification. We use cross-entropy loss for semantic segmentation and an adversarial loss for domain alignment in the generator, while the discriminator is trained using binary cross-entropy. Training is performed for 50,000 iterations with a batch size of 8. We use SGD with a learning rate of 0.005 and momentum 0.9 for the generator, and Adam with a learning rate of 0.0001 for the discriminator. The adversarial loss is scaled by a weight factor $\lambda = 0.01$. Detailed implementation is provided in Supplementary Section S2.

\subsection{Metrics} \label{section4_metrics}
All model evaluations are conducted on real-world validation data. Synthetic datasets are used only during training. This ensures that reported metrics reflect true generalization on real imagery.

To quantify segmentation performance, we use Intersection over Union (IoU), Overall Accuracy (OA), and F1 Score. IoU measures the overlap between predicted and ground truth segmentation, making it the primary metric for evaluating model performance in building detection. Accuracy provides an overall evaluation of correctly predicted pixels, while F1 captures the balance between precision and recall, which is particularly useful for imbalanced datasets.

The metrics are computed from the confusion matrix components: True Positives (TP), False Positives (FP), True Negatives (TN), and False Negatives (FN), as defined below:

\begin{equation}
\mathrm{IoU} = \frac{TP}{TP + FP + FN}
\end{equation}

\begin{equation}
\mathrm{OA} = \frac{TP + TN}{TP + TN + FP + FN}
\end{equation}

\begin{equation}
\mathrm{Precision} = \frac{TP}{TP + FP}
\end{equation}

\begin{equation}
\mathrm{Recall} = \frac{TP}{TP + FN}
\end{equation}

\begin{equation}
F_1 = 2 \times \frac{\mathrm{Precision} \times \mathrm{Recall}}{\mathrm{Precision} + \mathrm{Recall}}
\end{equation}

\subsection{Real-world dataset v.s. Synthetic augmented dataset} \label{section4_exp_real_vs_synthetic}

\subsubsection{Overall Performance Across the Region} \label{subsubsection4_overall}

Our experiments are designed to systematically evaluate the generalization benefits of geo-typical synthetic data across varied domain conditions. Each experiment provides a controlled evaluation setting, allowing us to assess performance under different source-target combinations. Collectively, the results highlight the consistent advantage of our proposed synthetic augmentation strategy in reducing domain gaps across geographically diverse urban areas. Specifically, we examine the impact of incorporating geo-typical synthetic data into building segmentation models by comparing models trained on real-world datasets alone with those augmented by synthetic data. 

A partial visual comparison is provided in Figure~\ref{fig:fig_results}, and performance gains are quantitatively summarized using IoU as the primary metric in Table~\ref{tab:performance_synthetic_data}. For instance, when the TIR dataset is supplemented with Cbus\_{s} data, there is a significant increase of approximately 0.12 in IoU, rising from 0.35 to 0.47. In various locations, when the target domain is CHI and source data from ATX, TIR, and DSTL are used, training that incorporates CHI\_{s} data has led to significant improvements. The results increased as follows: from 0.44 to 0.52 (+0.08) in ATX, from 0.35 to 0.45 (+0.10) in TIR, and from 0.06 to 0.34 (+0.28) in DSTL.

\begin{table}[H]
    \centering
    \begin{tabular}{c|c|c|c|c|c}
        \hline
        \textbf{Source} & \textbf{Target} & \textbf{wo/} & \textbf{w/ (Ours)} & \textbf{SynRS3D} & \textbf{SyntheWorld} \\
        \hline
        \hline
        Cbus & Cbus & \textbf{0.76}& 0.75 (-0.01) & 0.38 (-0.38) & 0.51 (-0.24)\\
        ATX & Cbus & 0.46& \textbf{0.52}  (+0.06)& 0.42 (-0.04) & 0.33 (-0.13)\\
        TIR & Cbus & 0.35& \textbf{0.47} (+0.12) & 0.28 (-0.07) & 0.15 (-0.20)\\
        DSTL & Cbus & 0.07 & \textbf{0.36}  (+0.29) & 0.03 (-0.04) & 0.01 (-0.06)\\
        \hline
        CHI & CHI & \textbf{0.81}& \textbf{0.81} (+0.00) & 0.49 (-0.32) & 0.53 (-0.28)\\
        ATX & CHI & 0.44         & \textbf{0.52} (+0.08) & 0.43 (-0.01) & 0.43 (-0.01)\\
        TIR & CHI & 0.35         & \textbf{0.45} (+0.10) & 0.23 (-0.12) & 0.37 (+0.02)\\
        DSTL & CHI & 0.06        & \textbf{0.34} (+0.28) & 0.11 (+0.05) & 0.02 (-0.04)\\
        \hline
    \end{tabular}
    \textit{\\*Note: 'wo/' indicates training without the inclusion of synthetic data, while 'w/' denotes training with target location corresponding geo-typical synthetic data (Cbus\_{s} with Cubs, CHI\_{s} with CHI). Values in parentheses indicate the IoU difference between training with and without synthetic data.}
    \caption{Performance of synthetic augmented data compared with only real-world data for building IoU}
    \label{tab:performance_synthetic_data}
\end{table}

These improvements can be attributed to several factors. First, geo-typical synthetic data based on the target domain's city layout introduces variability and diversity, helping the model learn features that reduce the domain gap. This alignment with the target domain's urban morphology ensures the model is trained on data resembling the target environment, enhancing segmentation accuracy. Additionally, the synthetic view generation phase improves generalization by exposing the model to various lighting conditions, reducing overfitting.

In Table \ref{tab:performance_synthetic_data}, the ATX dataset, with Cbus\_{s} augmentation, shows an improvement in IoU from 0.46 to 0.52, while coupling with SynRS3D and SyntheWorld yields an IoU of 0.42 (-0.04) and 0.33 (-0.13), respectively. The TIR dataset, when trained with Cbus\_{s} data, exhibits a substantial IoU increase from 0.35 to 0.47 (+0.12), but shows a drop to 0.28 (-0.07) and 0.15 (-0.20) with SynRS3D and SyntheWorld. These results suggest that geo-typical synthetic data improves model performance, while non-geo-typical data may introduce larger domain gaps, especially when the source or augmentation data does not match the target environment.

The drop in performance in Table \ref{tab:performance_synthetic_data}, when Cbus is used as both the source and target (IoU decreasing from 0.76 to 0.75 with Cbus\_{s} can be attributed to several factors. These include the mismatch between synthetic and real-world data characteristics, backpropagation during training can fall into a local optimum, complexities introduced by the synthetic datasets, challenges in generalizing across a blended feature space of real and synthetic data, and potential variations in data quality between the synthetic and real-world datasets.


\subsubsection{Performance in commercial and residential areas} \label{subsubsection4_commercial_residential}

Training with TIR, DSTL as sources, coupled with geo-typical synthetic data, has achieved promising results in detecting both residential and commercial buildings. For target area, two residential areas were chosen for comparison. These regions are located in Columbus, USA, and Chicago, USA, and include dense and sparse residential buildings. Additionally, two commercial areas were chosen, encompassing high-rise commercial buildings in Chicago and mid-rise commercial mixed academic buildings in Columbus. These four areas cover a total of 6.02 km\textsuperscript{2}.

As shown in Table \ref{tab:commercialResidentialAreas}. The integration of geo-typical synthetic data improved IoU by average 0.23 for commercial and 0.15 for residential areas. The IoU improvements in the target domain for the CHI commercial area increased by 0.28, representing the highest increase among all evaluated regions. The unique spatial grid-like layout of Chicago's commercial area is uncommon in the source domain. By incorporating geo-typical synthetic data that closely aligns with the target domain's features, the model achieved substantial performance gains, highlighting the critical role of geo-typical synthetic data in reducing the domain gap.

\begin{table}[H]
    \centering
    \begin{tabular}{c|c|c|c|c|c}
        \hline
        \multirow{2}{*}{\textbf{Source}} & \multirow{2}{*}{\textbf{Target}} & \multicolumn{2}{c|}{\textbf{Commercial}} &\multicolumn{2}{c}{\textbf{Residential}} \\
         & & \textbf{wo/} & \textbf{w/} &\textbf{wo/} & \textbf{w/}  \\
        \hline
        \hline
        \multirow{2}{*}{TIR} & Cbus  & 0.24     &  \textbf{0.45} (+0.21)   & 0.21   &  \textbf{0.39} (+0.18)   \\
                             & CHI   & 0.28     & \textbf{0.56}  (+0.28)  & 0.47   &  \textbf{0.53} (+0.06)   \\
        \hline
        \multirow{2}{*}{DSTL} & Cbus & 0.09     & \textbf{0.31} (+0.22)   & 0.01   &  \textbf{0.21} (+0.20)    \\
                              & CHI  & 0.30     & \textbf{0.51} (+0.21)  & 0.28   &  \textbf{0.44} (+0.16)    \\
        \hline
    \end{tabular}
    \textit{\\*Note: 'wo/' indicates training without the inclusion of synthetic data (Cbus\_{s}), while 'w/' denotes training with geo-typical synthetic data (Cbus\_{s}). Values in parentheses indicate the IoU difference between training with and without synthetic data.}
    \caption{IoU comparison for commercial and residential building detection}
    \label{tab:commercialResidentialAreas}
\end{table}

\subsection{Ablation study}\label{section4_ablation_study}
\subsubsection{Adversarial domain adaptation} \label{subsubsection4_domain_adaptation}

We conducted an ablation study to evaluate the impact of ADA on building segmentation performance across various sites, as summarized in Table \ref{tab:domain_adaptation_impact} using the IoU metric. The results indicate a slight IoU decrease for Cbus (-0.03) when ADA is applied, suggesting that ADA may have limited effectiveness when the source and target domains are already well aligned. In such cases, the adversarial loss introduces unnecessary regularization, it continues to apply gradient signals even when feature distributions are already similar, leading to feature drift rather than meaningful alignment. This disrupts the natural class boundaries learned during segmentation and degrades performance by forcing the model to unnecessarily adapt, resulting in suboptimal decision boundaries and reduced segmentation accuracy. Conversely, the inclusion of ADA significantly improves model performance in geo-typical synthetic data augmented. ATX shows a notable IoU increase of 0.21, while TIR and DSTL exhibit improvements of 0.06 and 0.22, respectively.

These findings highlight the advantages of proposed ADA in enhancing model adaptability and robustness, particularly in scenarios where the target environment differs from the training data. While a variety of alternative DA methods continue to emerge, our primary objective is to demonstrate the clear effectiveness of ADA combined with our geo-typical synthetic data approach. Conducting extensive comparative benchmarking with many other ADA variant is beyond our intended scope and research focus.

\begin{table}[H]
    \centering
    \begin{tabular}{c|c|c|c|c}
        \hline
        \textbf{Source} & \textbf{Syn. Data}  & \textbf{Target} & \textbf{wo/ADA} & \textbf{w/ADA} \\
        \hline
        \hline
        Cbus & Cbus\_{s} & Cbus & \textbf{0.78} & 0.75 (-0.03)\\
        ATX  & Cbus\_{s} & Cbus & 0.31& \textbf{0.52} (+0.21)\\
        TIR & Cbus\_{s} & Cbus & 0.41& \textbf{0.47} (+0.06)\\
        DSTL & Cbus\_{s} & Cbus & 0.14& \textbf{0.36} (+0.22)\\
        \hline
    \end{tabular}
    \textit{\\*Note: 'w/ADA' indicates training with adversarial domain adaptation (ADA), 'wo/ADA' without it. Values in parentheses indicate the IoU difference between training with and without ADA.}
    \caption{Comparative IoU analysis of model performance with and without DA}
    \label{tab:domain_adaptation_impact}
\end{table}

\subsubsection{Backbone selection} \label{subsubsection4_Backbones}

To evaluate the impact of the backbone network on segmentation performance, we conduct an ablation study comparing the original backbone (ResNet101) used in CLAN and HRNet+OCR under identical training conditions. Table \ref{tab:backbone_ablation} summarizes the IoU scores obtained when using different source domains (Cbus, ATX, TIR, and DSTL) while keeping the synthetic data fixed as Cbus\_{s} and targeting the Cbus domain.

The results (shown in Table \ref{tab:backbone_ablation}) indicate that HRNet+OCR consistently outperforms ResNet101 across all source domains. For instance, when the source is TIR, HRNet+OCR improves IoU from 0.28 to 0.46, a significant gain of +0.18. On average, HRNet+OCR achieves an IoU improvement ranging from +0.09 to +0.18, highlighting its stronger spatial representation capabilities.

The superior performance of HRNet+OCR compared to ResNet101 can be attributed to two key architectural advantages. First, HRNet maintains high-resolution representations throughout the network by fusing multi-scale features, which is particularly beneficial for semantic segmentation tasks requiring fine-grained boundary information. In contrast, ResNet101 progressively downsamples the input, which can lead to loss of spatial details critical for accurate pixel-wise predictions. Second, the OCR module enhances feature representations by capturing the relationship between each pixel and the object region it belongs to. This allows the model to aggregate more meaningful context at the object level, leading to improved segmentation performance, especially in cases with domain shifts or when transferring from synthetic to real-world scenes.

\begin{table}[H]
    \centering
    \begin{tabular}{c|c|c|c|c}
        \hline
        \textbf{Source} & \textbf{Syn.Data} & \textbf{Target} & \textbf{ResNet101} & \textbf{HRNet+OCR} \\
        \hline
        \hline
        Cbus & Cbus\_{s} & Cbus & 0.65 & \textbf{0.75} (+0.09) \\
        ATX & Cbus\_{s} & Cbus & 0.41& \textbf{0.51}  (+0.10)\\
        TIR & Cbus\_{s} & Cbus & 0.28& \textbf{0.46}  (+0.18)\\
        DSTL & Cbus\_{s} & Cbus & 0.21& \textbf{0.35}  (+0.14)\\
        \hline
    \end{tabular}
    \textit{\\*Note: Values in parentheses indicate the IoU difference between training with and without ADA.}
    \caption{Comparative IoU analysis of model performance with ResNet101 and with HRNet+OCR}
    \label{tab:backbone_ablation}
\end{table}

\subsubsection{Architecture comparison}
\label{subsubsection4_architecture_comparison}

To further examine the generalization capability enabled by our approach, we train UNet \cite{cao2022swin} and SegFormer \cite{xie2021segformer} independently on the same geo-typical synthetic dataset (Cbus\_s), without ADA. Both models use an 80/20 train-validation split with identical data patches. UNet is trained for 20 epochs, and SegFormer for 100 epochs, with model selection based on the best validation loss.

As shown in Table~\ref{tab:architecture_ablation}, both models underperform relative to our proposed CLAN + HRNet+OCR approach. We attribute this to several factors. UNet, while lightweight and efficient, lacks the representational and contextual aggregation mechanisms required for robust domain generalization. On the other hand, SegFormer has significantly higher model capacity with parameter count, when trained with limited data, likely leads to overfitting. Despite early stopping based on validation loss, we observed that SegFormer started to memorize training patterns, degrading its ability to generalize to target domain. This suggests that in low-resource training scenarios, high-capacity transformer-based models may require additional regularization or larger-scale synthetic data to realize their potential.

In contrast, our pipeline combines geo-typical synthetic supervision with adversarial domain alignment, enabling more stable generalization across domains, even with moderate training data size.

\begin{table}[H]
    \centering
    \begin{tabular}{c|c|c|c|c|c}
        \hline
        \textbf{Source} & \textbf{Syn.Data} & \textbf{Target} & \textbf{Ours} & \textbf{UNet} & \textbf{Segformer} \\
        \hline
        \hline
        Cbus & Cbus\_{s} & Cbus & \textbf{0.75} & 0.74 (-0.01) & 0.57 (-0.18) \\
        ATX  & Cbus\_{s} & Cbus & \textbf{0.51} & 0.27 (-0.24) & 0.17 (-0.34) \\
        TIR  & Cbus\_{s} & Cbus & \textbf{0.46} & 0.30 (-0.16) & 0.14 (-0.31) \\
        DSTL & Cbus\_{s} & Cbus & \textbf{0.35} & 0.22 (-0.13) & 0.16 (-0.19) \\
        \hline
    \end{tabular}
    \textit{\\*Note: All models are trained and evaluated on the same data splits. Values in parentheses indicate the IoU difference relative to our method (CLAN + HRNet + OCR).}
    \caption{IoU performance comparison of different architectures (UNet, SegFormer) trained on the same geo-typical synthetic data without domain adaptation.}
    \label{tab:architecture_ablation}
\end{table}

\subsubsection{Texture domain randomization} \label{subsubsection4_domain_randomization}

In this analysis, we investigate the impact of introducing texture DR on building segmentation model performance across four distinct sites. Table \ref{tab:domain_ramdomization_impact} presents a comparative assessment of models trained with and without texture DR, using IoU as the evaluation metric.

The results from Cbus, with minimal variation (a 0.01 decrease in IoU with texture DR), suggest that texture DR has a limited effect in environments closely resembling the training data. However, the more significant improvements at ATX (0.18 increase), TIR (0.20 increase), and DSTL (0.14 increase) with the introduction of texture DR indicate its substantial benefit in scenarios where the real-world dataset is different from the model's initial training conditions.

These findings underscore the effectiveness of texture DR in enhancing the model robustness, particularly in environments that differ considerably from the original training dataset. The marked improvements in IoU at ATX, TIR, and DSTL demonstrate that texture DR can significantly contribute to the model's ability to generalize across varied and challenging scenarios, thus proving to be a vital tool in scenarios where dataset diversity is limited or the target environment is not well-represented in the training data.

\begin{table}[H]
    \centering
    \begin{tabular}{c|c|c|c}
        \hline
        \textbf{Source} & \textbf{Target} & \textbf{wo/DR} & \textbf{w/DR}  \\
        \hline
        \hline
        Cbus  & Cbus & \textbf{0.76} & 0.75 (-0.01)\\
        ATX & Cbus & 0.34& \textbf{0.52} (+0.18)\\
        TIR  & Cbus & 0.27& \textbf{0.47} (+0.20)\\
        DSTL  & Cbus & 0.22& \textbf{0.36} (+0.14)\\
        \hline
    \end{tabular}
    \textit{\\*Note: 'w/DR' denotes training data has DR (Cbus\_{s}), while 'wo/DR' indicates training data has no DR (Cbus\_{sw}). Values in parentheses indicate the IoU difference between training with and without DR.}
    \caption{Comparative IoU analysis of model performance with and without texture domain randomization (DR)}
    \label{tab:domain_ramdomization_impact}
\end{table}

\subsubsection{City layout} \label{subsubsection4_citylayout}

A numerical analysis detailing the influence of Cbus\_{s} and CHI\_{s} on the model's performance reveals distinct patterns (shown in Table \ref{tab:City_performance_synthetic}). When comparing the two synthetic datasets, Cbus\_{s} leads to better performance than CHI\_{s}. This is evident in the experiment where Cbus, ATX, TIR, DSTL are the source. The model's IoU trained with Cbus\_{s} is consistently better compared to CHI\_{s}. For instance, despite the decrease in introducing the synthetic data in the Cbus to Cbus experiment. the IoU for augmented Cbus\_{s} (0.75) is almost the same compare to augmented CHI\_{s} (0.75). Further evidence of this pattern is seen in other source datasets. For TIR, an increase in IoU is observed when augmented with Cbus\_{s}, reaching 0.47, while with CHI\_{s}, the IoU is slightly lower at 0.42. Similarly, for ATX, the IoU increases to 0.52 with Cbus\_{s} and to 0.51 with CHI\_{s}. The DSTL dataset shows a pronounced difference: the IoU substantially increases to 0.36 with Cbus\_{s}, compared to a lesser increase to 0.32 with CHI\_{s}. 

These variations underscore that the specific characteristics of synthetic data have an impact on model performance. Cbus\_{s}, due to its features being more aligned with the Cbus dataset, assists the model in better capturing the details of the target environment. In contrast, CHI\_{s}, appears to introduce elements that are less effective for the model trained on Cbus data. This implies that the choice of synthetic data needs careful consideration, taking into account how well it matches the target dataset's attributes. For more details please refer to Supplementary Section S3.
\begin{table}[H]
    \centering
    \begin{tabular}{c|c|c|c}
        \hline
        \textbf{Source} & \textbf{Target} & \textbf{Non-aligned}& \textbf{Aligned}\\
        \hline
        \hline
        Cbus & Cbus & \textbf{0.75} & \textbf{0.75} (+0.00) \\
        ATX & Cbus & 0.51& \textbf{0.52} (+0.01) \\
        TIR & Cbus & 0.42& \textbf{0.47} (+0.05) \\
        DSTL & Cbus & 0.32& \textbf{0.36} (+0.04) \\
        \hline
    \end{tabular}
    
    \textit{\\*Note: 'Non-aligned' refers to training with CHI\_{s} (layout not aligned with the target domain), while 'Aligned' uses Cbus\_{s} (layout aligned). Values in parentheses indicate the IoU difference between training with non-aligned and with aligned city layout of synthetic data.}
    \caption{Comparative IoU performance of aligned and non-aligned city layout of synthetic data}
    \label{tab:City_performance_synthetic}
\end{table}

\section{Conclusion}\label{sec:conclusion}
In this study, we propose the use of ADA (CLAN incorporating HRNet + OCR) coupled our geo-typical synthetic data to improve building segmentation in VHR satellite images. Our experiments show that with our approach significantly boosts the performance of deep learning models, achieving up to a 12\% median improvement and, in some cases, performance gains of up to 29\%. This underscores the effectiveness of coupling geo-typical synthetic datasets for ADA, especially when real-world data is insufficient to capture target domain diversity.

Utilizing procedural modeling and physics-based rendering based on OpenStreetMap street network, we generate geo-typical synthetic datasets that closely mimic the target domain's city layout, enhancing model generalization across diverse geographical regions. We also provide these synthetic datasets to support further research and development.

Additionally, our synthetic coupling approach is adaptable to other categories, such as roads and vegetation, making it versatile for various remote sensing applications. Future work will refine the geo-typical synthetic data generation process, integrate more urban / suburb characteristics, and evaluate the method's applicability to other remote sensing tasks, with a focus on scaling to larger and more diverse datasets.

\section*{Acknowledgement}
The authors would like to thank Dr. Bowen Wen for helping render the datasets.

\bibliographystyle{IEEEtran}
\bibliography{references}

\begin{IEEEbiography}[{\includegraphics[width=1in,height=1.25in,clip,keepaspectratio]{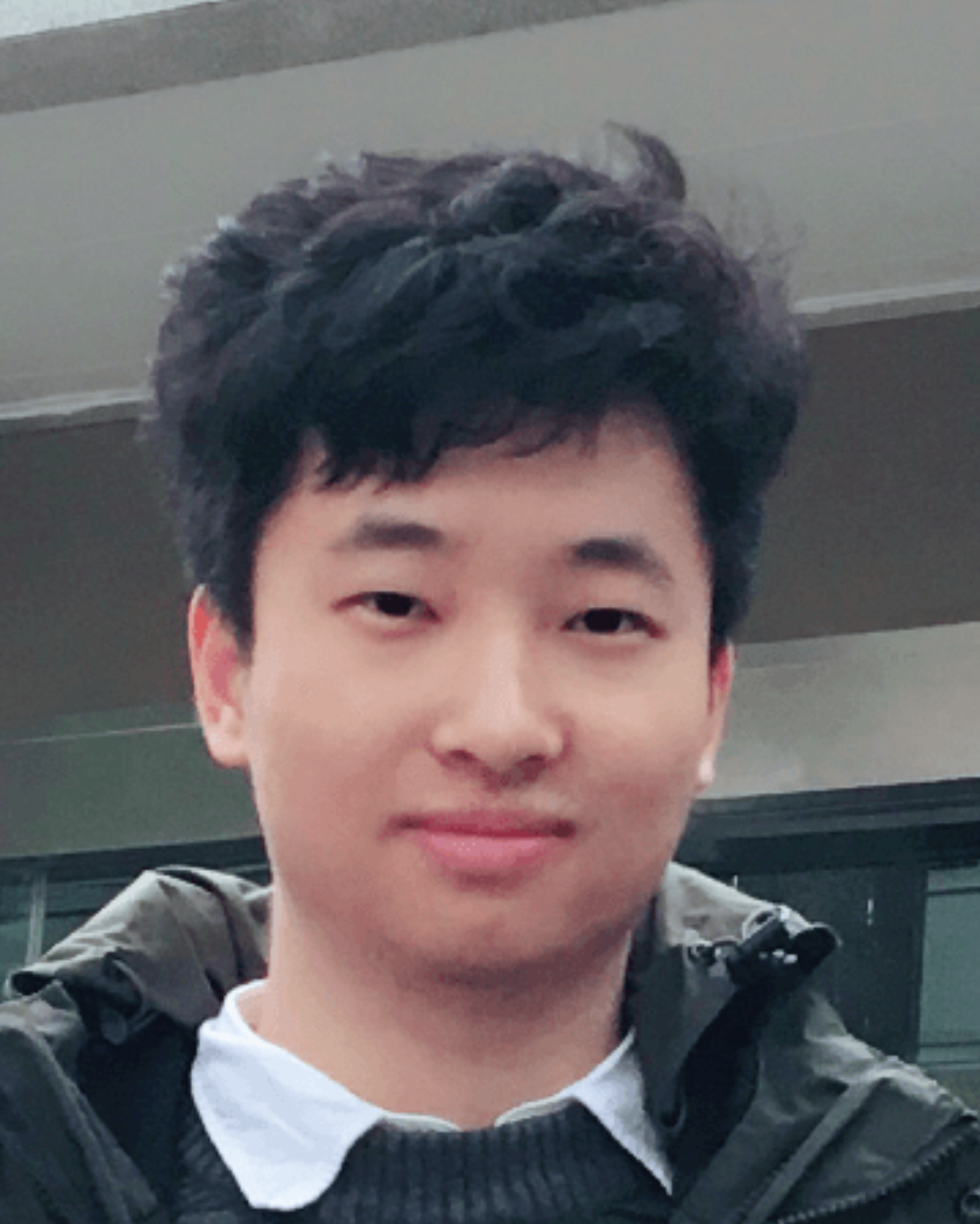}}]{Shuang Song} received the B.S. degree in geographic information science and cartography from Guangzhou University, Guangzhou, China, in 2014, and the M.S. and Ph.D. degree in photogrammetry and remote sensing from Wuhan University, Wuhan and geoinformation and geodetic engineering from The Ohio State University, Columbus in 2024. He is currently a research scientist at the Department of Translational Data Analytics Institute at The Ohio State University, Columbus. His research interests include Remote Sensing, Photogrammetry and Machine Learning.
\end{IEEEbiography}

\begin{IEEEbiography}[{\includegraphics[width=1in,height=1.25in,clip,keepaspectratio]{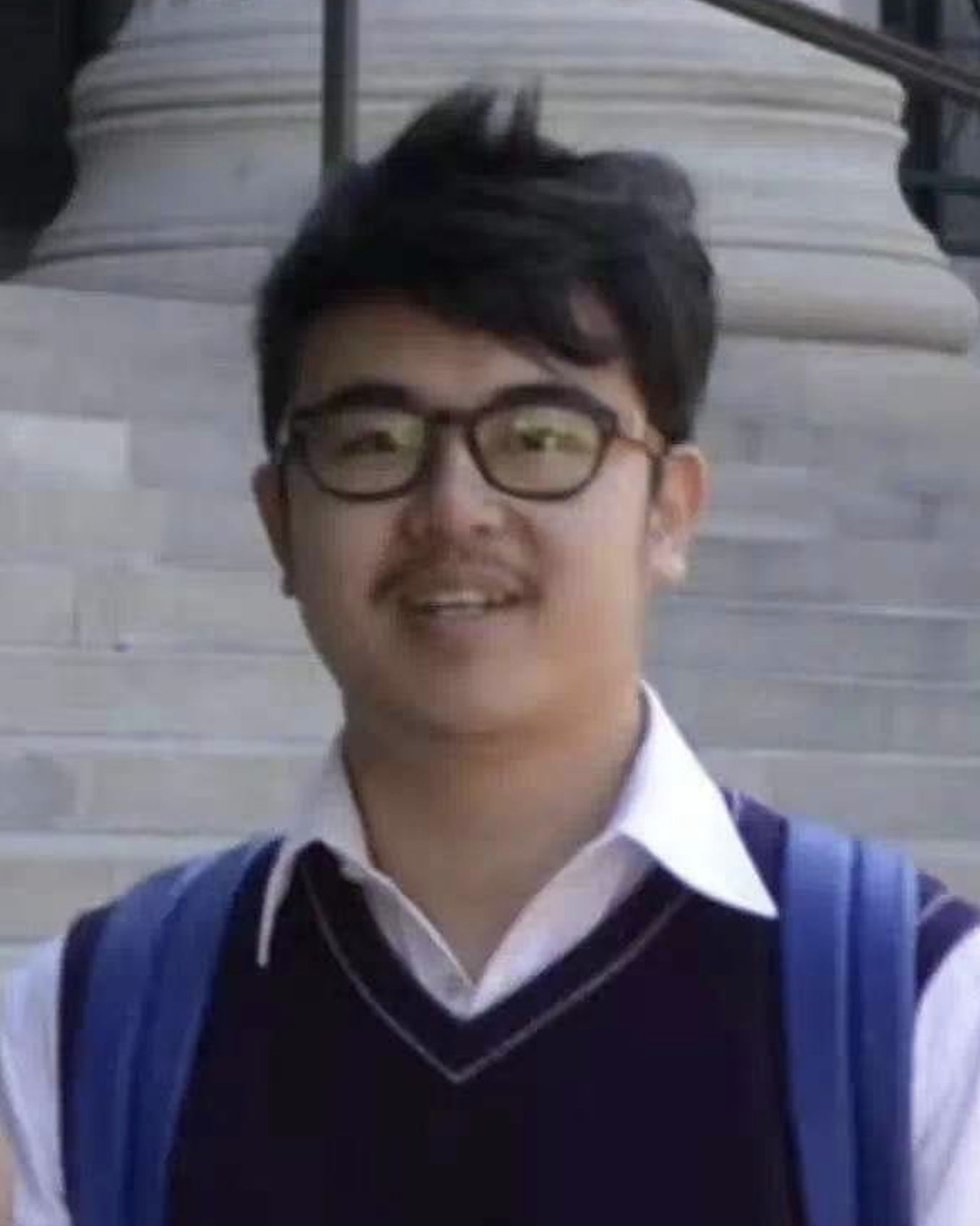}}]{Yang Tang} received the B.S. degree in computer science from Wrocław University of Science and Technology, Wroclaw, Poland, in 2012, and the M.S. degree in digital media engineering from Denmark Technical University, Copenhagen and Tokyo University, Tokyo in 2014. He is currently working toward the Ph.D. degree in the Department of Civil, Environmental and Geodetic Engineering at The Ohio State University, Columbus. His research interests include remote sensing image classification, change detection, and hyperspectral imaging.
\end{IEEEbiography}

\begin{IEEEbiography}[{\includegraphics[width=1in,height=1.25in,clip,keepaspectratio]{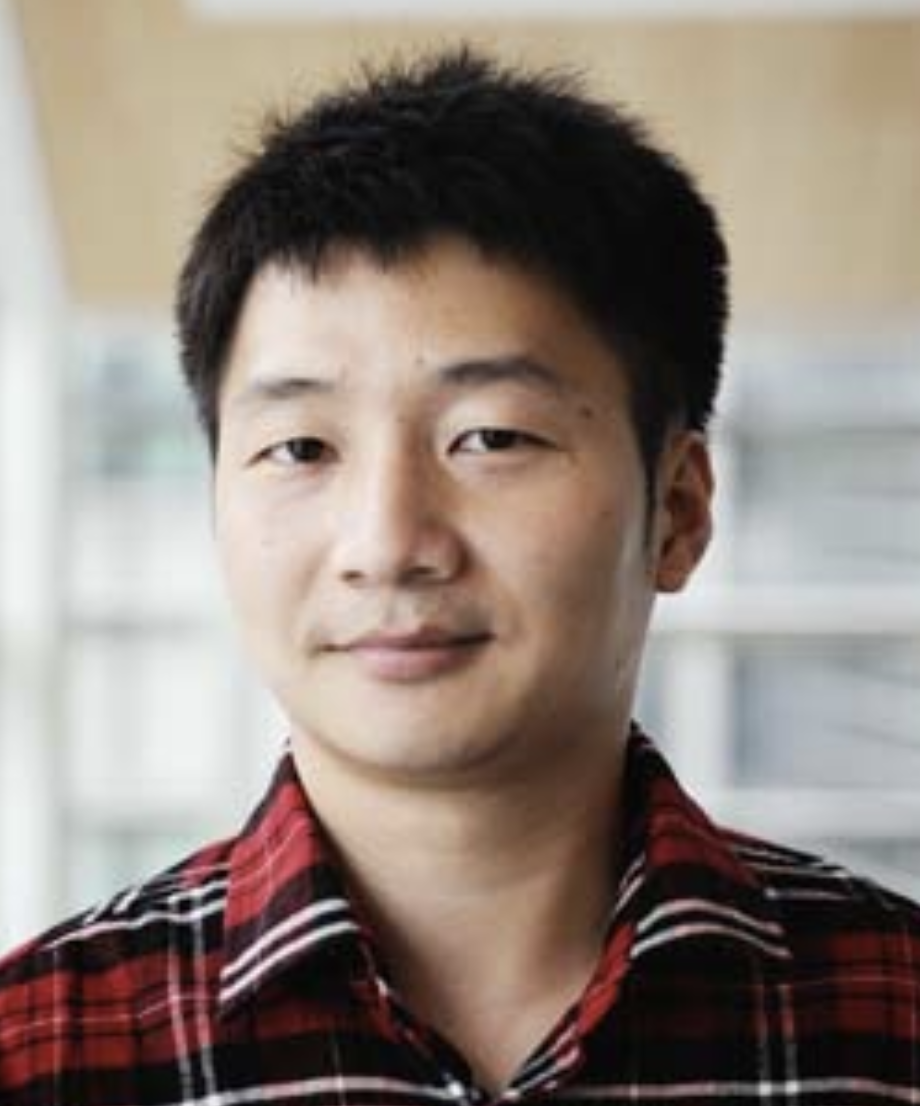}}]{Rongjun Qin} received the B.S. degree in computational mathematics from Wuhan University, Wuhan, China, in 2009, and the M.S. and Ph.D. degree in photogrammetry and remote sensing from Wuhan University and ETH, Zurich in 2011 and 2015 respectively. He is currently a
faculty member of the Department of Civil, Environmental and Geodetic Engineering, Department of Electrical and Computer Engineering at The Ohio State University, Columbus. His research interests include photogrammetric 3D reconstruction, remote sensing image classification, UAV image processing, image dense matching and change detection. His research seeks computational solutions to various geometric and interpretation
problems in an urban context using imaging sensors such as aerial/UAV imagery, LiDAR, and satellite multispectral/hyperspectral images. Prof. Qin is the author of RSP (RPC stereo processor) and MSP (multi-stereo processor) used for reconstructing 3D information from 2D images with high quality. Prof. Qin is an associate editor for the Photogrammetric Engineering and Remote Sensing journal. He is also chairing the working group “Satellite Constellation for Remote Sensing” of the International Society for Photogrammetry and Remote Sensing Commission. His awards include the first prize in Mathematical Modeling Contest and several other prominent scholarship awards. Prof. Qin serves as a reviewer for more than 15 international journals in the field of photogrammetry and remote sensing. 
\end{IEEEbiography}
\vfill


\end{document}